\documentclass[conference]{IEEEtran}
\IEEEoverridecommandlockouts                              

\usepackage{graphicx}
\usepackage[letterpaper, left=0.75in, right=0.75in, bottom=0.8in, top=0.75in]{geometry}

\usepackage{amsmath} 
\usepackage{amssymb}  
\usepackage{mathrsfs}
\usepackage{multirow}

\usepackage{todonotes}
\usepackage{enumerate}
\usepackage{float}
\usepackage{caption}
\usepackage{subcaption}
\usepackage{cite}
\usepackage{tabularx}
\usepackage[ruled,vlined,linesnumbered]{algorithm2e}
\usepackage{bm}
\usepackage{xcolor}
\usepackage{mathrsfs}

\title{\LARGE \bf
Decentralized Gaussian Process Classification and an Application in Subsea Robotics \thanks{This work was supported by the Seale Coastal Zone Observatory and the Office of Naval Research via grants N00014-23-1-2345, N00014-24-1-2267, and N00014-25-1-2224.} \thanks{Yifei Gao, Hans J. He, and Daniel Stilwell are with the Bradley Department of Electrical and Computer Engineering, Virginia Tech, Blacksburg, VA, USA, James McMahon is with the Acoustics Division, Code 7130, US Naval Research Laboratory, Washington, D.C. }
}

\author{\IEEEauthorblockN{Yifei Gao}
\IEEEauthorblockA{\textit{Virginia Tech}\\
yifeig22@vt.edu}
\and

\IEEEauthorblockN{Hans J. He}
\IEEEauthorblockA{\textit{Virginia Tech}\\
	hjh2bs@vt.edu}

\and
\IEEEauthorblockN{Daniel J. Stilwell}
\IEEEauthorblockA{\textit{Virginia Tech}\\
	stilwell@vt.edu}

\and
\IEEEauthorblockN{James McMahon}
\IEEEauthorblockA{\textit{US Naval Research Laboratory}\\ Acoustics Division, Code 7130 \\
    james.mcmahon@nrl.navy.mil}

}

\begin{document}
\maketitle
\thispagestyle{empty}
\pagestyle{empty}

\begin{abstract}
    Teams of cooperating autonomous underwater vehicles (AUVs) rely on acoustic communication for coordination, yet this communication medium is constrained by limited range, multi-path effects, and low bandwidth.  One way to address the uncertainty associated with acoustic communication is to learn the communication environment in real-time. 
 We address the challenge of a team of robots building a map of the probability of communication success from one location to another in real-time. This is a decentralized classification problem -- communication events are either successful or unsuccessful -- where AUVs share a subset of their communication measurements to build the map.  The main contribution of this work is a rigorously derived data sharing policy that selects measurements to be shared among AUVs. We experimentally validate our proposed sharing policy using real acoustic communication data collected from teams of Virginia Tech 690 AUVs, demonstrating its effectiveness in underwater environments.
\end{abstract}

\section{INTRODUCTION}

We address challenges imposed by teams of cooperating autonomous underwater vehicles that seek to coordinate their actions underwater.  Communication between AUVs is necessary to support coordination, but the acoustic communication channel available to AUVs is extremely low-bandwidth often unreliable \cite{stojanovic2009underwater} \cite{horner2013data}.  To maximize the ability of AUVs to collaborate in real-time, we seek a method for the AUVs to cooperatively build a map that can be used to predict the probability of communication success.  We address this goal as a Gaussian process classification problem that seeks to estimate the probability of successful or unsuccessful communication from a transmitting AUV at a specific location to a receiving AUV at another location. Implementing Gaussian process classification across a team of agents that have only very low bandwidth communication channels between agents requires novel advances in decentralized Gaussian process classification, which is the principal contribution of our work. 

In practice, a team of AUVs can be pre-programmed with the maximum range at which they can successfully communicate.  This range is necessarily conservative.  Instead, we seek an approach to a team of AUVs developing a map of the probability of communication success in real-time based on their current experience attempting to communicate in the environment.  A conservative range can be used as a prior, and the possibility of communicating at longer ranges can be learned in real-time.  In past work, we have investigated Gaussian process classification in a centralized manner for this application \cite{gao2024prediction}.  In this work, we explicitly address the challenge of decentralized Gaussian process classification (GP) so that the team of AUVs can learn communication success maps in real-time.  Toward this goal, we develop fundamental advances in GP classification that can be applied to other application.

In the literature, most of the work including our prior work \cite{horner2013data, zhao2018efficient,quattrini2020multi, 8647266, clark2022propeml, b12, b13} use Gaussian process regression to predict received signal strength (RSS) or use a manifold learning algorithm to predict signal to noise ratio (SNR) \cite{kazemi2020snr}. In \cite{gao2024prediction}, we address the application using GP classification, which allows us to explicitly learn and predict the variable of interest, which is the probability of communication success, rather than a proxy variable, such as SNR. Our prior work \cite{gao2024prediction} employs Gaussian process classification to generate a communication map among two AUVs in a centralized manner suitable for post-processing.  We also take advantage of the properties of P\'{o}lya-Gamma random variables in \cite{polson2013bayesianinferencelogisticmodels} to make prior conjugate, which is not otherwise the case for GP classification.
\begin{figure}[t!]
  \centering
  \includegraphics[width=2.0in]{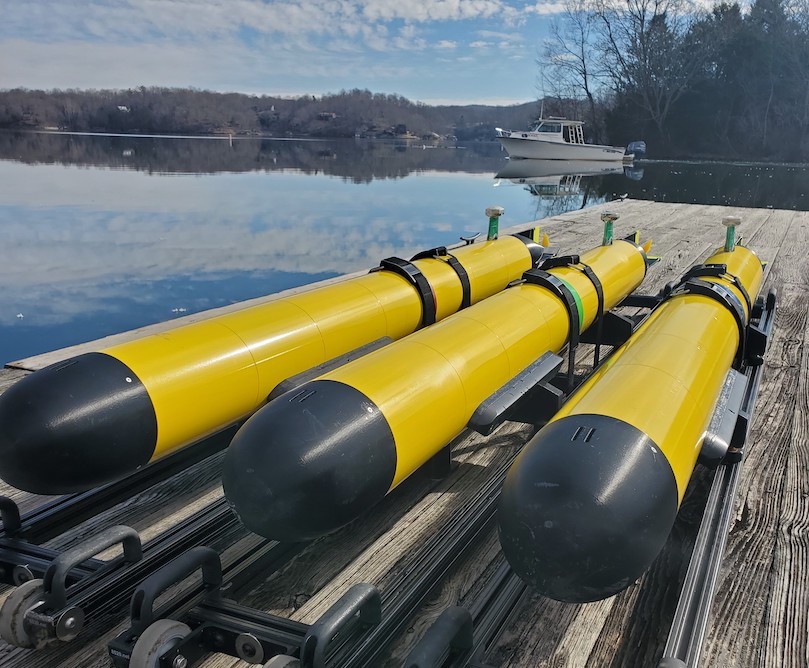}
  \caption{Virginia Tech 690 AUVs}
\label{AUV_figure}
\end{figure}

We use sparse Gaussian process classification to represent the communication map. Due to the low-bandwidth communication channel, the map should be represented with as few measurements of virtual measurements as possible. 

Gaussian process classification is used in a variety of applications that range from image classification  \cite{bazi2009gaussian} to  collision prediction for robot motion planning \cite{munoz2023collisiongp}. GP classification is widely used in  active learning for label prediction \cite{zhao2021efficient}. A computation challenge of GP classification arises from the prior being Bernoulli and therefore non-conjugate.  Recent work that addresses this challenge with clever use of  P\'{o}lya-Gamma random variables \cite{wenzel2018efficientgaussianprocessclassification} \cite{achituve2021personalized} \cite{snell2020bayesian}, which we adopt in our work.  Our approach to sparse GP classification, which builds upon the seminal works \cite{wenzel2018efficientgaussianprocessclassification} \cite{achituve2021personalized}\cite{burt2020convergence}.  The use of sparse GP classification is because we seek a minimal representation of the environment due to the low bandwidth communication channel available underwater.

\noindent The contributions of this work are
\begin{itemize}
    \item We provide a rigorously derived policy selecting data to share among agents for decentralized sparse GP classification under limited communication bandwidth.
    \item For an application in underwater communication, We show empirically the efficacy of our data sharing policy.  We conduct numerical experiments using real data from field trials with teams of small AUVs that communicate acoustically. 
\end{itemize}

The remainder of the paper is as follows. In Section II we review sparse GP classification with P\'{o}lya-Gamma random variables. In Section III, we derive an upper bound for the KL divergence between the estimated posterior and true posterior which we use to establish a data sharing policy. In Section IV we illustrate and evaluate our results on real communication data set acquired by a team of AUVs.

\section{Preliminaries}
In this section, we briefly review sparse Gaussian process binary classification with P\'{o}lya-Gamma random variables. For further details are found in  \cite{machine_learning},\cite{wenzel2018efficientgaussianprocessclassification}, \cite{polson2013bayesianinferencelogisticmodels}, \cite{titsias}, \cite{titsias2014variational}.
\subsection{Gaussian Process Binary Classification}
In standard Gaussian process (GP) classification, we consider learning a latent function $f : \mathbb{R}^{d} \rightarrow \mathbb{R}$ from measurements $\{\textbf{x}, y\}$ where $\textbf{x} \in \mathbb{R}^{d}$ and binary label $y \in \{0,1\}$. We assume this latent function is sampled from a Gaussian process $\mathcal{GP}(\mu(\textbf{x}), k(\textbf{x}, \textbf{x}'))$ where $\mu$ denotes the latent mean function and $k : \mathcal{X} \times \mathcal{X} \rightarrow \mathbb{R}$ is a kernel function. The posterior closed-form formula for Gaussian process regression (i.e., learning $f$) is covered in \cite{machine_learning}. 

For GP binary classification, we want to learn the probability distribution 
\begin{equation}
p(y_{*} = 1| \textbf{y}, \textbf{x}_{*}) = \int p(y_{*} = 1| f_{*})p(f_{*} | \textbf{y}, \textbf{x}_{*}) df_{*}
\label{eq:posterior_fstar}
\end{equation}
where $\textbf{y}$ is the set of all training measurement labels and $\textbf{x}_{*}$ is the test input. The conditional likelihood $p(y_{n} | f_{n})$ is defined as a logistic function
\begin{equation*}
p(y_{n} | f_{n}) = \left(\frac{e^{f_{n}}}{1 + e^{f_{n}}}\right)^{y_{n}}\left( 1 - \frac{e^{f_{n}}}{1 + e^{f_{n}}}\right)^{(1 - y_{n})} \label{eq2}
\end{equation*}
which is non-Gaussian. The logistic function maps the range of $f$ ($\mathbb{R}$) to the interval $[0,1]$, inducing a valid probabilistic interpretation. We want to predict the probability of a successful communication event given any test location $\textbf{x}_{*}$. The latent posterior $p(f_{*} | \textbf{y}, \textbf{x}_{*})$ on \eqref{eq:posterior_fstar} is non-Gaussian since the likelihood is non-conjugate. However, the logistic likelihood can be made conjugate by introducing a P\'{o}lya-Gamma random variable \cite{polson2013bayesianinferencelogisticmodels}. 

\subsection{Gaussian process classification with P\'{o}lya-Gamma (PG) random variable}
As proposed in \cite{polson2013bayesianinferencelogisticmodels}, P\'{o}lya-Gamma random variable  $w \sim PG(b,c)$ with probability density function
\begin{equation*}
    p(w | b,c) = \frac{1}{2 \pi^{2}} \sum_{k = 1}^{\infty} \frac{g_{k}}{(k - \frac{1}{2})^{2} + (\frac{c}{2 \pi})^{2}}
\end{equation*}
where $b > 0$, $c \in \mathbb{R}$ and $g_{k} \sim Ga(b,1)$ are independent Gamma random variables. We assume an independent observation model conditioned with PG random variables $\boldsymbol{w}$,
\begin{equation*}
    p(\textbf{y} | \boldsymbol{w}, \textbf{f}) = \prod_{n = 1}^{N} p(y_{n} | w_{n}, f_{n})
\end{equation*}
with
\begin{equation*}
    p(y_{n} | w_{n}, f_{n}) = \frac{1}{2} \exp \bigg\{ - \frac{w_{n}}{2} \bigg( f_{n}^{2} - 2f_{n} \frac{\kappa_{n}}{w_{n}}\bigg)\bigg\}
\end{equation*}
where $w_{n} \sim PG(1, c_{n})$ and $\kappa_{n} = y_{n} - 1/2$. Based on the Laplace transform property of P\'{o}lya-Gamma random variable \cite{polson2013bayesianinferencelogisticmodels}, the joint conditional likelihood with PG variables is proportional to a Gaussian distribution
\begin{equation*}
    p(\textbf{y} | \boldsymbol{w}, \textbf{f}) \propto \mathcal{N}(\boldsymbol{\Omega}^{-1} \boldsymbol{\kappa} | \textbf{f}, \boldsymbol{\Omega}^{-1})
\end{equation*}
where $\boldsymbol{\Omega} = diag(\boldsymbol{w})$ is a diagonal matrix of P\'{o}lya-Gamma variables and $\boldsymbol{\kappa} = [\kappa_{1},..., \kappa_{N}]^{T} \in \mathbb{R}^{N}$. The prior of latent function is assumed $p(\textbf{f}) = \mathcal{N}(\textbf{f} | \textbf{0}, \textbf{K}_{NN})$ where $\textbf{K}_{NN}$ is the Gram matrix taking total measurements $\textbf{X}$ as input. Conditioned on $\mathbf{y}$ and the PG random variables $\boldsymbol{w}$, the posterior over the latent function values is also Gaussian
\begin{equation*}
    p(\textbf{f} | \textbf{y}, \boldsymbol{w}) = \mathcal{N}(\textbf{f} | \boldsymbol{\mu}, \boldsymbol{\Sigma})
\end{equation*}
with
\begin{equation*}
\boldsymbol{\mu} = \boldsymbol{\Sigma} \boldsymbol{\kappa}, \qquad 
        \boldsymbol{\Sigma} = (\textbf{K}_{NN}^{-1} + \boldsymbol{\Omega})^{-1}
\end{equation*}
\subsection{Sparse Gaussian process with P\'{o}lya-Gamma Variables}
Computing the posterior distribution $p(\textbf{f} | \textbf{y}, \boldsymbol{w})$ requires the 
$\mathcal{O}(N^{3})$ procedure of inverting the Gram matrix $\textbf{K}_{NN}$. The author of \cite{titsias} proposes sparse Gaussian process to reduce computational complexity to $\mathcal{O}(M^{3})$, introducing $M$ inducing points to summarize total data set with size $N$ where $M \ll N$. Denote the inducing locations as $\textbf{Z} = [\textbf{z}_{1},...,\textbf{z}_{M}]^{T}$, $\textbf{z}_{i} \in \mathbb{R}^{d}$, and let $\textbf{f}_{M} = [f_{1}(\textbf{z}_{1}),...,f_{M}(\textbf{z}_{M})]^{T}$ be the inducing output vector. The conditional prior on latent function values $\textbf{f}$ given $\textbf{f}_{M}$ is
\begin{equation*}
    p(\textbf{f} | \textbf{f}_{M}) = \mathcal{N}(\textbf{f} | \textbf{K}_{\textbf{X}, \textbf{Z}}\textbf{K}_{\textbf{Z}, \textbf{Z}}^{-1}\textbf{f}_{M}, \textbf{K}_{\textbf{X}, \textbf{X}} - \textbf{K}_{\textbf{X}, \textbf{Z}}\textbf{K}_{\textbf{Z}, \textbf{Z}}^{-1}\textbf{K}_{\textbf{X}, \textbf{Z}}^{T})
\end{equation*}
where kernel matrix $\textbf{K}_{\textbf{X}, \textbf{Z}} \in \mathbb{R}^{N \times M}$ and $\textbf{K}_{\textbf{Z}, \textbf{Z}} \in \mathbb{R}^{M \times M}$. Note that now we only need to invert kernel matrix $\textbf{K}_{\textbf{Z}, \textbf{Z}}$. In \cite{titsias}, the author shows that we can estimate the latent posterior as $p(\textbf{f} | \textbf{y}) \approx \int p(\textbf{f} | \textbf{f}_{M}) q(\textbf{f}_{M}) d \textbf{f}_{M}$ using the variational distribution  $q(\textbf{f}_{M})$. However, in standard GP classification, the variational distribution for inducing outputs does not have a closed-form expression. In next section, we show that while conditioning on P\'{o}lya-Gamma random variables $\boldsymbol{\omega}$, we can compute a 
 closed-form expression for $q(\textbf{f}_{M} | \boldsymbol{w}) = \mathcal{N}(\textbf{f}_{M} | \boldsymbol{\mu}_{\boldsymbol{w}}, \boldsymbol{\Sigma}_{\boldsymbol{w}})$ where $\boldsymbol{\mu}_{\boldsymbol{w}}$ and $\boldsymbol{\Sigma}_{\boldsymbol{w}}$ are the mean vector and covariance matrix conditioned on $\boldsymbol{w}$. This yields a closed-form formula for the latent posterior conditioned on P\'{o}lya-Gamma variables $p(\textbf{f} | \textbf{y}, \boldsymbol{w})$,
\begin{equation*}
    p(\textbf{f} | \textbf{y}, \boldsymbol{w}) \approx \int p(\textbf{f} | \textbf{f}_{M}) q(\textbf{f}_{M} | \boldsymbol{w}) d \textbf{f}_{M} = \mathcal{N}(\textbf{f} | \textbf{m}, \textbf{S})
\end{equation*}
where 
\begin{align*}
    \textbf{m} &= \textbf{K}_{\textbf{X}, \textbf{Z}}\textbf{K}_{\textbf{Z}, \textbf{Z}}^{-1}\boldsymbol{\mu}_{\boldsymbol{w}} \\
    \textbf{S} &= \textbf{K}_{\textbf{X},\textbf{X}} - \textbf{K}_{\textbf{X}, \textbf{Z}}\textbf{K}_{\textbf{Z},\textbf{Z}}^{-1}\textbf{K}_{\textbf{X},\textbf{Z}}^{T} + \textbf{K}_{\textbf{X},\textbf{Z}}\textbf{K}_{\textbf{Z},\textbf{Z}}^{-1}\boldsymbol{\Sigma}_{\boldsymbol{w}}\textbf{K}_{\textbf{Z},\textbf{Z}}^{-1}\textbf{K}_{\textbf{X},\textbf{Z}}^{T} 
\end{align*}

\section{Selection of inducing points}
In this section, we first derive the upper bound for KL divergence between the estimated and true posterior inspired from \cite{burt2020convergence}. Then we show this upper bound can be used as to select inducing points to share. To see the detailed derivation of the Kullback-Leibler(KL) divergence upper bound in Gaussian process regression, please refer to \cite{burt2020convergence}.
\subsection{Lower and Upper bound for Log Marginal likelihood}
We assume that we are given a set of known measurements of communication success acquired by an agent (an AUV, in our case) that consist of locations  $\textbf{X} = [\textbf{x}_{1},...,\textbf{x}_{N}]^{T}$ and a corresponding labels $y_{n} \in \{0,1\}$. By assuming independent observations, the conditional likelihood can be written 
\begin{equation*}
    p(\textbf{y} | \boldsymbol{w}, \textbf{f}) = \prod_{n = 1}^{N} p(y_{n} | w_{n},f_{n}) = C\mathcal{N}(\boldsymbol{\Omega}^{-1} \boldsymbol{\kappa} | \textbf{f}, \boldsymbol{\Omega}^{-1})
\end{equation*}
where $C$ is a unknown constant that arises from the joint conditional likelihood being proportional to a Gaussian distribution. The marginal likelihood conditioned on P\'{o}lya-Gamma random variables $\boldsymbol{w}$ is
\begin{equation}
    p(\textbf{y} | \boldsymbol{w}) = \int p(\textbf{y} | \boldsymbol{w}, \textbf{f}) p(\textbf{f}) d\textbf{f} = C\mathcal{N}(\boldsymbol{\Omega}^{-1} \boldsymbol{\kappa} | \textbf{0}, \boldsymbol{\Omega}^{-1} + \textbf{K}_{NN})\label{marginal_likelihood_eq}
\end{equation} 
Similarly, we  follow the derivation in \cite{titsias} and \cite{titsias2014variational} to derive a lower and upper bound for $\log p(\textbf{y} | \boldsymbol{w})$ which is used to compute upper bound of KL divergence between estimated and true posterior. We compute directly 
\begin{equation} \label{eq.l_lower}
	\begin{split}
			\log p(\textbf{y} | \boldsymbol{w}) &\geq \log C - \frac{N}{2} \log (2 \pi) \\
			&- \frac{1}{2} \log |\boldsymbol{\Sigma}| - \frac{1}{2} (\boldsymbol{\Omega}^{-1} \boldsymbol{\kappa})^{T}\boldsymbol{\Sigma}^{-1} (\boldsymbol{\Omega}^{-1} \boldsymbol{\kappa}) \\
            &- \frac{1}{2} tr(\boldsymbol{\Omega} \tilde{\textbf{K}}) = \mathcal{L}_{lower}
	\end{split}
\end{equation}
where $\boldsymbol{\Sigma} = \boldsymbol{\Omega}^{-1} + \textbf{K}_{NM} \textbf{K}_{MM}^{-1} \textbf{K}_{MN}$ and $\tilde{\textbf{K}} = \textbf{K}_{NN} - \textbf{K}_{NM}\textbf{K}_{MM}^{-1} \textbf{K}_{MN}$. Note that $\mathcal{L}_{lower}$ lower bounds the log marginal likelihood, and if  $\textbf{X} = \textbf{Z}$, then $tr(\boldsymbol{\Omega} \tilde{\textbf{K}}) = 0$ and $\mathcal{L}_{lower} = \log p(\textbf{y} | \boldsymbol{w})$. We can show that while conditioning on P\'{o}lya-Gamma variables $\boldsymbol{w}$, the optimal closed-form distribution for inducing variables $q(\textbf{f}_{M} | \boldsymbol{w})$ can be written as
\begin{equation}
	q(\textbf{f}_{M} | \boldsymbol{w}) = \mathcal{N}(\textbf{f}_{M} | \textbf{K}_{MM} \tilde{\boldsymbol{\Sigma}}^{-1} \textbf{K}_{MN} \boldsymbol{\kappa}, \textbf{K}_{MM} \tilde{\boldsymbol{\Sigma}}^{-1} \textbf{K}_{MM}) \label{optimal_closed_form_eq}
\end{equation}
where $\tilde{\boldsymbol{\Sigma}} = \textbf{K}_{MM} + \textbf{K}_{MN} \boldsymbol{\Omega} \textbf{K}_{NM}$ and $\kappa_{n} = y_{n} - \frac{1}{2}$ is one of the entries in vector $\boldsymbol{\kappa}$ with $y_{n} \in \{0,1\}$. 

Following the derivation in \cite{titsias2014variational}, we can write an upper bound for log marginal likelihood conditioned on P\'{o}lya-Gamma variable, 
\begin{equation}\label{eq.l_upper}
	\begin{split}
		\log p(\textbf{y} | \boldsymbol{w}, \textbf{X}) &\leq \log C - \frac{N}{2} \log(2\pi) \\
        &- \frac{1}{2} \log |\boldsymbol{\Omega}^{-1} + \textbf{K}_{NM} \textbf{K}_{MM}^{-1} \textbf{K}_{MN}| \\
		&- \frac{1}{2} \textbf{c}^{T} (\boldsymbol{\Omega}^{-1} + p\textbf{I} + \textbf{K}_{NM} \textbf{K}_{MM}^{-1} \textbf{K}_{MN})^{-1} \textbf{c}\\
        &= \mathcal{L}_{upper}
	\end{split}
\end{equation}
where $\textbf{c} = \boldsymbol{\Omega}^{-1} \boldsymbol{\kappa}$ and $p = tr(\tilde{\textbf{K}})$. If  $\textbf{X} = \textbf{Z}$, then we again arrive at $\mathcal{L}_{upper} = \log p(\textbf{y} | \boldsymbol{w})$. 

\subsection{KL Divergence Upper bound} \label{sec.KL_upper}
Following the analysis in \cite{burt2020convergence}, an upper bound is derived for KL divergence between the estimated posterior $q(\textbf{f}, \textbf{f}_{M}) = p(\textbf{f} | \textbf{f}_{M}) q(\textbf{f}_{M} | \boldsymbol{w})$ and true posterior $p(\textbf{f}, \textbf{f}_{M} | \boldsymbol{w}, \textbf{y}) = p(\textbf{f} | \textbf{f}_{M}) p(\textbf{f}_{M} | \boldsymbol{w}, \textbf{y})$ which is denoted as $KL[q(\textbf{f}, \textbf{f}_{M}) || p(\textbf{f}, \textbf{f}_{M} | \boldsymbol{w}, \textbf{y})] \triangleq KL[Q_{\boldsymbol{w}} || P_{\boldsymbol{w}}]$.  We don't know the true inducing posterior conditioned on P\'{o}lya-Gamma variables $\boldsymbol{w}$ denoted as $p(\textbf{f}_{M} | \boldsymbol{w}, \textbf{y})$ but we can use $q(\textbf{f}_{M} | \boldsymbol{w})$ to approximate it. This KL divergence quantifies how close we are to the true posterior conditioned on $\boldsymbol{w}$. Let $\log$ marginal likelihood conditioned on $\boldsymbol{w}$ denoted as $\mathcal{L} = \log p(\textbf{y} | \boldsymbol{w})$. From \eqref{eq.l_lower} and \eqref{eq.l_upper}, we can have $\mathcal{L}_{lower} \leq \mathcal{L} \leq \mathcal{L}_{upper}$ and we can also directly show that $\log p(\textbf{y} | \boldsymbol{w}) = KL[Q_{\boldsymbol{w}} || P_{\boldsymbol{w}}] + \mathcal{L}_{lower}$ from which we write the upper bound $KL[Q_{\boldsymbol{w}} || P_{\boldsymbol{w}}] \leq \mathcal{L}_{upper} - \mathcal{L}_{lower}$. Note that the constant $C$ in the joint conditional likelihood $p(\textbf{y} | \boldsymbol{w}, \textbf{f})$ is eliminated in the  difference between upper and lower bound of the $\log$ marginal likelihood $\log p(\textbf{y} | \boldsymbol{w})$. Therefore, we can write
\begin{equation*}
    \begin{split}
        KL[Q_{\boldsymbol{w}} || P_{\boldsymbol{w}}] \leq \frac{1}{2} tr(\tilde{\textbf{K}}) \lambda_{\max}(\boldsymbol{\Omega}) + \frac{1}{2\lambda_{\min}(\boldsymbol{\Omega})} ||\boldsymbol{\kappa}||_{2}^{2}
    \end{split}
\end{equation*}
where $\tilde{\textbf{K}} = \textbf{{K}}_{NN} - \textbf{K}_{NM} \textbf{K}_{MM}^{-1} \textbf{K}_{MN}$. $\lambda_{\max}(\boldsymbol{\Omega})$ and $\lambda_{\min}(\boldsymbol{\Omega})$ are maximum and minimum eigenvalue of diagonal matrix $\boldsymbol{\Omega}$ hence largest and smallest corresponding entries in the diagonal. Since we assume that we are given P\'{o}lya-Gamma random variables $\boldsymbol{w}$ to form $\boldsymbol{\Omega}$, only $tr(\tilde{\textbf{K}})$ and $||\boldsymbol{\kappa}||_{2}^{2}$ affect the KL divergence upper bound. We cannot control $||\boldsymbol{\kappa}||_{2}^{2}$ since it scales with size  $N$ of the current data set. 



Selecting inducing points which minimize $tr(\tilde{\textbf{K}})$ is non-trivial but \cite{burt2020convergence} \cite{titsias2014variational} \cite{belabbas2009spectral} \cite{anari2016monte} propose determinantal point process and ridge leverage scores to select $m \leq M$ number of inducing points in an acceptable amount of computation time. However, in this paper, we are only interested in selecting small number of inducing points where $m \leq 2$ due to limited communication bandwidth.


\subsection{Communication Among Multiple agents}
In this section we examine precisely what information should be shared among agents. We assume that each agent is within its own locality region and denote local measurements obtained by agent $i$ be $\{\textbf{X}_{i}, \textbf{y}_{i}\}$. We assume each agent's locality region is independent, and that each agent selects inducing points directly from their measurement locations $\textbf{X}_{i}$.  Each agent's set of inducing points  $\textbf{Z}_{i}$ are therefore independent with respect to those of other agents. Suppose agent $i$ chooses $m_{i}$ inducing points from its current inducing set $M_{i}$ in its locality region which minimizes $tr(\tilde{\textbf{K}}_{i})$,
\begin{equation*}
    tr(\tilde{\textbf{K}}_{i}) = tr(\textbf{K}_{N_{i} N_{i}} - \textbf{K}_{N_{i} m_{i}} \textbf{K}_{m_{i} m_{i}}^{-1} \textbf{K}_{m_{i} N_{i}})
\end{equation*}
where $N_{i}$ denotes the total number of local data set size obtained from agent $i$ in the locality region. After selecting $m_{i}$ number of inducing points, agent $i$ computes the variational distribution $q(\textbf{f}_{m_{i}} | \boldsymbol{w}_{i}) = \mathcal{N}(\textbf{f}_{m_{i}} | \boldsymbol{\mu}_{m_{i}}, \boldsymbol{\Sigma}_{m_{i}})$ using \eqref{optimal_closed_form_eq} with $\textbf{K}_{MM}$ and $\textbf{K}_{NM}$ replaced by $\textbf{K}_{m_{i} m_{i}}$ and $\textbf{K}_{N_{i} m_{i}}$. The information package which each agent needs to share is $\bigg\{ \textbf{Z}_{m_{i}}, q(\textbf{f}_{m_{i}} | \boldsymbol{w}_{i})\bigg\}$,
where $\textbf{Z}_{m_{i}}$ denotes $m_{i}$ inducing locations selected to minimize $tr(\tilde{\textbf{K}}_{i})$. Based on the upper limit of the KL divergence derived Section \ref{sec.KL_upper}, agent $i$ selects $m_{i}$ inducing points and computes $q(\textbf{f}_{m_{i}} | \boldsymbol{w}_{i})$, which  minimizes the upper bound of KL divergence in its locality region. 

\subsection{Prediction}
For the case of $n$ agents, each agent summarizes its locality region with $q(\textbf{f}_{m_{i}} | \boldsymbol{w}_{i})$, which it shares with other agents. We denote the complete set of inducing variables as $\textbf{f}_{M} = [\textbf{f}_{m_{1}},..., \textbf{f}_{m_{n}}]$ where $\textbf{f}_{m_{i}} = f(\textbf{Z}_{m_{i}})$ are the latent function vectors taking inducing locations $\textbf{Z}_{m_{i}}$ as input. Then we can write the joint variational distribution after each agent $i$ receives information packages $\bigg\{\textbf{Z}_{m_{j}}, q(\textbf{f}_{m_{j}} | \boldsymbol{w}_{j})\bigg\}$ where $i \neq j$ as $q(\textbf{f}_{M} | \boldsymbol{w}) = \mathcal{N}(\textbf{f}_{M} | \boldsymbol{\mu}_{M}, \boldsymbol{\Sigma}_{M})$ where
\begin{equation}
    \boldsymbol{\mu}_{M} =
    \begin{bmatrix}
            \boldsymbol{\mu}_{m_{1}}\\
            \vdots\\
            \boldsymbol{\mu}_{m_{i}}\\
            \vdots\\
            \boldsymbol{\mu}_{m_{n}}
    \end{bmatrix}, 
    \boldsymbol{\Sigma}_{M} = 
    \begin{bmatrix}
        \boldsymbol{\Sigma}_{m_{1}} & \cdots & 0 & \cdots & 0\\
                                \vdots & \ddots & \vdots & \ddots & \vdots\\
                                0 & \cdots & \boldsymbol{\Sigma}_{m_{i}} &\cdots & 0\\
                                \vdots & \ddots & \vdots & \ddots & \vdots\\
                                0 & \cdots & 0 & \cdots & \boldsymbol{\Sigma}_{m_{n}}        
    \end{bmatrix}
\end{equation} \label{inducing_posterior}
where $\boldsymbol{\mu}_{M} \in \mathbb{R}^{M \times 1} $ and $\boldsymbol{\Sigma} \in \mathbb{R}^{M \times M}$ where $M = m_{1} + m_{2} +... + m_{n}$. 
Note this decentralized formulation (block-diagonal structure) is due to assuming independence between the locality region for each AUV, which we arrive at by noting that the effect of measurements outside of a locality region are vanishingly small. Therefore, the distribution for the value of the latent function at test location $\textbf{x}_{*}$ is 
\begin{equation*}
    \begin{split}
            &p(f_{*} | \textbf{y}, \boldsymbol{w}) \approx \int p(f_{*} | \textbf{f}_{M}) q(\textbf{f}_{M} | \boldsymbol{w}) d\textbf{f}_{M}\\
            &= \mathcal{N}(f_{*} | \textbf{k}_{n^{*}M} \textbf{K}_{MM}^{-1} \textbf{f}_{M}, k_{n_{*} n_{*}} - \textbf{k}_{n_{*} M} \textbf{K}_{MM}^{-1} \textbf{k}_{M n_{*}})
    \end{split}
\end{equation*}
where $p(f_{*} | \textbf{f}_{M})$ is the Gaussian conditional prior over $f_{*}$ which has a closed-form expression and  $\textbf{k}_{n_{*} M}$ is a vector in $\mathbb{R}^{1 \times M}$. Then the probability of success on test location $\textbf{x}_{*}$ given training data $\textbf{y}$ is approximated as
\begin{equation*}
    p(y_{*} = 1 | \textbf{y}, \boldsymbol{w}) \approx \int p(y_{*} | f_{*}) p(f_{*} | \textbf{y} , \boldsymbol{w}) df_{*}
\end{equation*}
which can be computed by 1-D Gaussian quadrature.

\section{Numerical Results}
We evaluate and illustrate our approach to decentralized sparse GP classification using data acquired from previous field trials with teams of Virginia Tech 690  autonomous underwater vehicles (see Figure \ref{AUV_figure}). The 690 AUV displaces approximately 43Kg and is 2.23 meters long. It communicates acoustically using the WHOI Micromdem 2, which operates at 25KHz. Experiments where performed with two AUVs in Claytor Lake, near Dublin, Virginia, USA, and with two and three AUVs in Massachusetts Bay near Boston, Massachusetts, USA.  The AUVs communicate using time-division multiple access communication; each AUV broadcasts to all other AUVs sequentially, with each broadcasting AUV assigned a sequence of known time-intervals.  Therefore if a receiving AUV does not receive a data packet during a time when it is expected, the occurrence of an unsuccessful communication event can be noted and recorded. 

The location of a communication event is represented by $\textbf{x} \in \mathbb{R}^{4}$, where the first two dimensions indicate the easting and northing coordinates of the broadcasting vehicle in the horizontal plane, and the last two dimensions contain the easting and northing coordinates of the receiving vehicle. For the data presented herein, all AUVs operated within a few meters of the same depth, so depth variations are not considered. Distance between AUVs is not explicitly included as a parameter since the relative positions of transmitting and receiving AUVs implicitly encode this information. To reduce computational complexity and enable straightforward visualization of communication events on 2D plots, environmental factors such as multipath effects and water current speed are not incorporated in this model. These factors will be addressed in future work.

We note that communication success measurements are obtained by all AUVs that receive or fail to receive a communication packet from a broadcasting AUV. The broadcasting AUV cannot assess success or failure.  The broadcasting AUV can be informed of a receiving AUV's communication measurements, but only via direct communication of the measurement from the receiving AUV.  Of course, the acoustic communication channel available to the AUVs is very low bandwidth, and we presume that all communication measurements cannot be shared among the team of AUVs. This motivates our development of a data-sharing policy (see Section III) for decentralized sparse Gaussian process classification.

We evaluate our decentralized sparse GP classification approach using three datasets as detailed in TABLE \ref{tab:auv_communication}. The column \textbf{Agent} indicates the specific AUV involved in each dataset. For each AUV, the  \emph{Unsuccessful Events} and \emph{Successful Events} columns represent the total number of failed and successful communication attempts when that particular AUV acted as the receiving vehicle. These measurements were obtained by the AUV, and are therefore stored locally onboard the AUV. For instance, in \textbf{Dataset 3},AUV 1 successfully received 119 data packets from AUV 2 or AUV 3, and it failed to receive 132 data packets from AUV 2 and AUV 3.


\begin{table*}[htbp]
\centering
\caption{AUV Communication Events Across Different Datasets}
\begin{tabular}{|c|c|c|c|c|}
\hline
\textbf{Dataset} & \textbf{Location} & \textbf{Agent} & \textbf{Unsuccessful Events} & \textbf{Successful Events} \\
\hline
\multirow{2}{*}{Dataset 1} & \multirow{2}{*}{Massachusetts Bay} & 1 & 248 & 198 \\
\cline{3-5}
 & & 2 & 219 & 227 \\
\hline
\multirow{2}{*}{Dataset 2} & \multirow{2}{*}{Claytor Lake} & 1 & 366 & 197 \\
\cline{3-5}
 & & 2 & 336 & 227 \\
\hline
\multirow{3}{*}{Dataset 3} & \multirow{3}{*}{Massachusetts Bay} & 1 & 132 & 119 \\
\cline{3-5}
 & & 2 & 105 & 130 \\
\cline{3-5}
 & & 3 & 103 & 109 \\
\hline
\end{tabular}
\label{tab:auv_communication}
\end{table*}

Using the data acquired from the field, we implement decentralized GP classification by sharing inducing points (i.e. virtual measurements in each broadcast data packet, and we choose which measurements to share using three different strategies. We select inducing points that minimize $tr(\tilde{\textbf{K}})$ as described on Section III, and we compare this approach to randomly selecting inducing points, and to selecting inducing points that maximize $tr(\tilde{\textbf{K}})$, which our analysis suggests should be the least useful inducing points. We address two cases: in the first we share one inducing point, and in the second we share two inducing points.  While an acoustic modem data packet may have room for more inducing points, the data packet is usually needed for communicating other missions-specific data.  Thus we believe our experiments with one and two inducing points is representative of real-world constraints.

We evaluate the classification prediction accuracy with the ration $\textbf{ACC} = \frac{N_{correct}}{N_{total}}$ where $N_{correct}$ is the number of correct prediction. $N_{total}$ is the total number of test points $\textbf{x}_{*}$.
If the true label is $y = 1$ , and prediction gives a probability larger than $0.5$, we treat it as a correct prediction and $y = 0$ is addressed similarly. We assess the certainty of the prediction using 
\begin{equation*}
    \textbf{NLL} = - \frac{1}{N_{total}}\sum_{i = 1}^{N_{total}} y_{i} \log(p_{i}) + (1 - y_{i}) \log(1 - p_{i})
\end{equation*}

For all three data sets, we use the squared exponential kernel function with unit signal variance, for any two inputs $\textbf{x}$ and $\textbf{x}'$
\begin{align*}
    k(\mathbf{x}, \mathbf{x}') = \exp \bigg(-\frac{\|\mathbf{x} - \mathbf{x}'\|^{2}}{2 l^{2}} \bigg)
\end{align*}

We initially standardize all communication data sets to make them have zero mean and unit variance. Then we use GPflow \cite{GPflow2017} to train on all standardized data sets and determine an optimal length-scale parameter $l=0.289$, which we then fix for all experiments across all agents. Note that real-time hyper-parameter learning is not explicitly addressed in this work. For each decentralized communication data set, we establish a reference radius $r$ based on the squared exponential kernel function with the fixed length-scale. This radius is defined:
\begin{equation*}
    r = l \cdot \sqrt{-2 \cdot \log(\epsilon)}
\end{equation*}
with $\epsilon$ tolerance where we have $k(\textbf{x}, \textbf{x}') < \epsilon$ if $||\textbf{x} - \textbf{x}'|| > r$. We use this radius to define a \emph{locality region} within each agent's local data set. Each region is centered at one of the local measurements stored in the agent's data set and encompasses all points within distance $r$ of this center. Note that locality region $R(r, \textbf{c})$ with center $\textbf{c} \in \mathbb{R}^{4}$ which we defined is a ball containing all locations $\textbf{l} \in \mathbb{R}^{4}$ which is 
\begin{equation*}
    R(r, \textbf{c}) = \{ \textbf{l} \in \mathbb{R}^{4} : ||\textbf{l} - \textbf{c}||_{2} \leq r\}
\end{equation*}
and $\textbf{c}$ is one of the measurement point obtained by AUV. Only data points inside this locality region are considered when training the sparse Gaussian process classification model.

Our approach assesses how effectively one or two inducing points can summarize the information provided by all measurements within a locality region. Before evaluating how our data sharing policy enables an AUV to make communication predictions in locality regions containing data points it has not yet encountered, we first conduct two local simulations by selecting inducing points within locality regions in each agent's local data set. If the selected good inducing points yield better predictions on test points within the locality, we can be confident that when an agent shares these inducing points with other agents, those received agents can make better predictions on this locality region even if they have not yet visited that region themselves.

For \textbf{Dataset 1} (see Table \ref{tab:auv_communication}) there are two AUVs.  We partition the data corresponding to which agent received or failed to receive a communication event and label partitioned data \textbf{Agent 1 local data set} and \textbf{Agent 2 local data set}. We then perform 100 random permutations for train-test splits on each agent's local dataset. In each permutation, 65 percent of the data is allocated for training and 35 percent for testing. For every permutation, we identify a small locality region defined by a radius $r=0.4$ with a fixed center $\textbf{c}$ and we select inducing points from this fixed locality region. It's important to note that since the communication data was previously standardized, this radius $r$ has no unit. We observe that increasing $r$ would necessitate more than one or two inducing points to adequately summarize the information within the locality region. In subsequent visualizations, we revert to the raw communication events data, where coordinates of each AUV are expressed in meters with northing and easting positions.

Since it is a random permutation, each locality region encompasses distinct training data points. Within each locality region, each agent selects $m = 1$ and $m = 2$ inducing points that effectively summarize the data within that locality. We classify inducing point selection strategies as follows: \emph{good} inducing point locations are selected by minimizing $tr(\tilde{\mathbf{K}})$, where we constrain selection to existing data point locations within the locality; \emph{bad} inducing point locations maximize $tr(\tilde{\mathbf{K}})$; and \emph{random} inducing point locations are randomly selected from data points $\mathbf{x}$ within the region.

After selecting the inducing locations $\mathbf{Z}_{m_i}$, each agent computes $q(\mathbf{f}_{m_i} | \boldsymbol{w}_i)$ based on equation \eqref{optimal_closed_form_eq} and uses this to predict the unseen 35 percent test data points within each identified locality region. We assume that $\boldsymbol{w}_{i}$ is available for each agent, we compute it via the Gibbs sampling technique described in \cite{achituve2021personalized}, which involves sampling from the Pólya-Gamma posterior $p(\boldsymbol{w}_i | \mathbf{f}_i)$
and $p(\mathbf{f}_i | \mathbf{y}_i, \boldsymbol{w}_i)$ for a specified number of iterations to produce a sampled $\boldsymbol{w}_i$.

Results for each agent selecting inducing points and evaluate on its test points inside locality region before sharing those inducing points are presented in Tables \ref{tab.local1} and \ref{tab.local2}. Note the values shown on the table are averaged over 100 since we have 100 random permutations. We observe that for \textbf{Agent 2}, selecting one or two good inducing points offers substantial prediction advantages over random or bad inducing points. This advantage is more pronounced compared to the \textbf{Agent 1}, as prediction performance is also strongly influenced by the distribution of local data within each locality region.

\begin{table}[h!] 
\caption{(\textbf{Dataset 1: Agent 1 local data set})}
\begin{center} \label{tab.local1}
\begin{tabular}{|c|c|c|c|c|c|} 
\hline
\textbf{SGPC-PG} & \textbf{ACC} & \textbf{NLL}\\
\hline

\hline\hline
\textbf{Good: select one / two} & \textbf{0.6498} / \textbf{0.6552} & 0.6685 / \textbf{0.6280}\\
\hline
\textbf{Random: select one / two} & 0.6458 / 0.6489 & 0.6546 / 0.6444\\
\hline
\textbf{Bad: select one / two} & 0.6427 / 0.6455 & 0.6558 / 0.6499\\
\hline
\end{tabular}
\end{center}
\end{table}

\begin{table}[h!] 
\caption{(\textbf{Dataset 1: Agent 2 local data set})}
\begin{center} \label{tab.local2}
\begin{tabular}{|c|c|c|c|c|c|} 
\hline
\textbf{SGPC-PG} & \textbf{ACC} & \textbf{NLL}\\
\hline

\hline\hline
\textbf{Good: select one / two} & \textbf{0.7152} / \textbf{0.7272} & \textbf{0.5902} / \textbf{0.5588}\\
\hline
\textbf{Random: select one / two} & 0.5747 / 0.6249 & 0.6706 / 0.6353\\
\hline
\textbf{Bad: select one / two} & 0.4838 / 0.4835 & 0.7004 / 0.7029\\
\hline
\end{tabular}
\end{center}
\end{table}

\begin{figure}[b]
        \centering
  \includegraphics[width=0.4\textwidth]{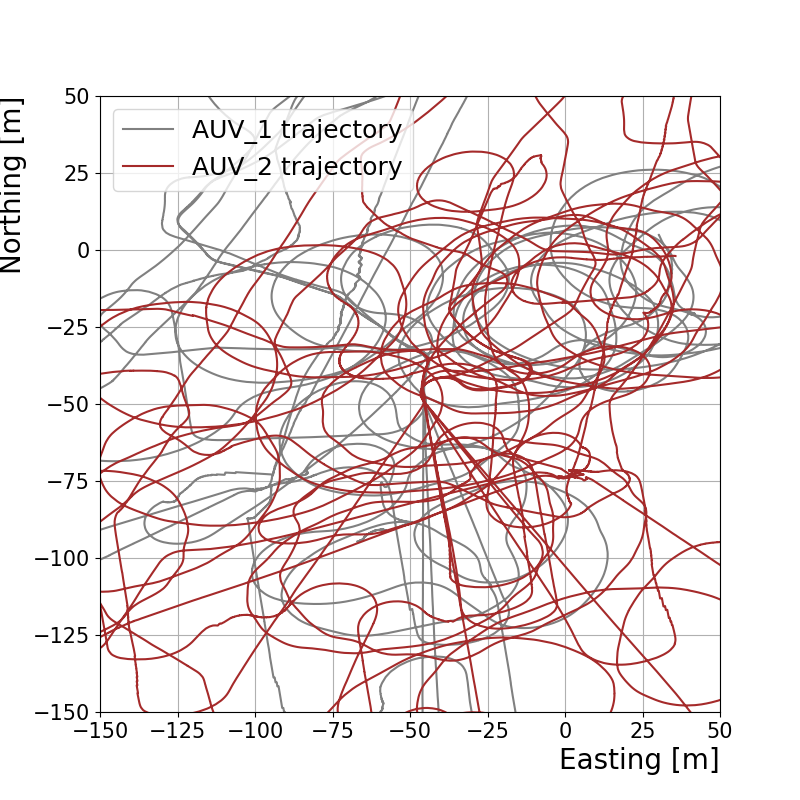}
\caption{Partial trajectories of two AUVs.}
\label{figure_local_inducing_picked}
\end{figure}

\begin{figure}[t!]\label{figure_local_inducing}
        \centering

    \centering
  \begin{subfigure}[b]{0.40\textwidth}
  \includegraphics[width=\textwidth]{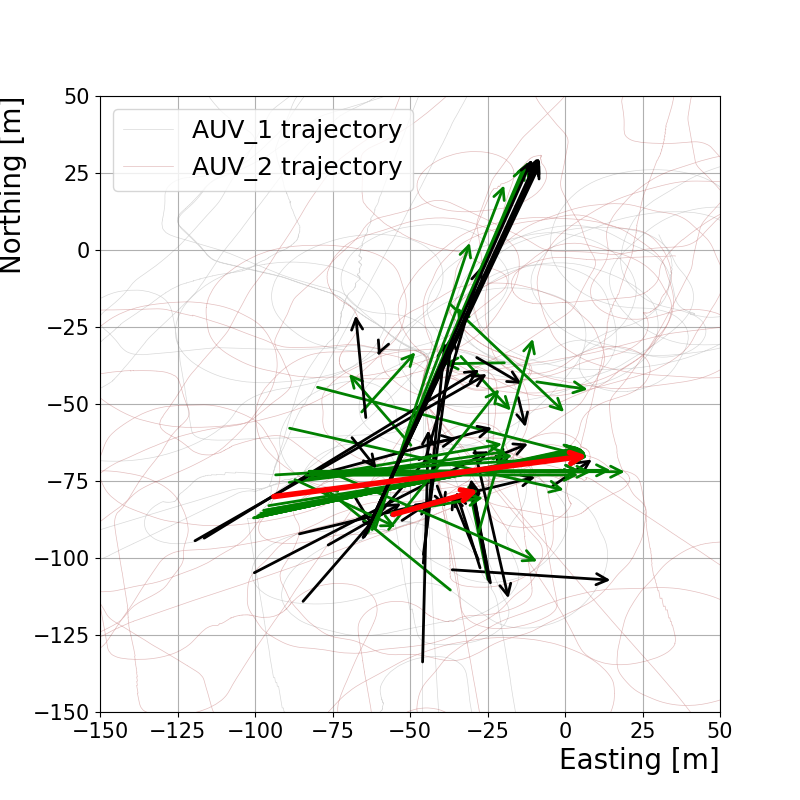}
  \end{subfigure}

  \centering
    \begin{subfigure}[b]{0.40\textwidth}
  \includegraphics[width=\textwidth]{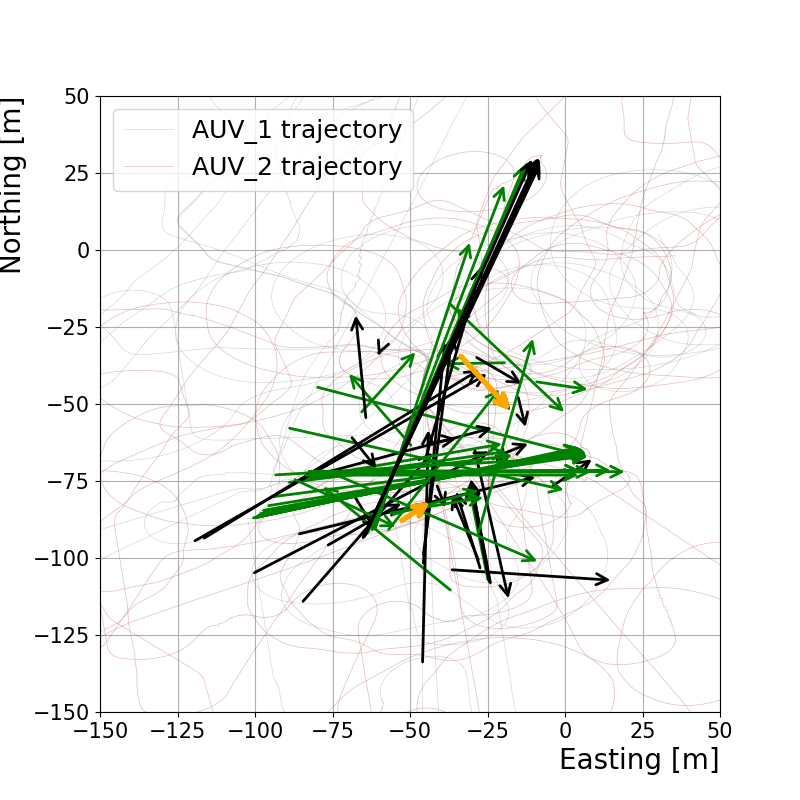}
  \end{subfigure}  

  \centering
    \begin{subfigure}[b]{0.40\textwidth}
  \includegraphics[width=\textwidth]{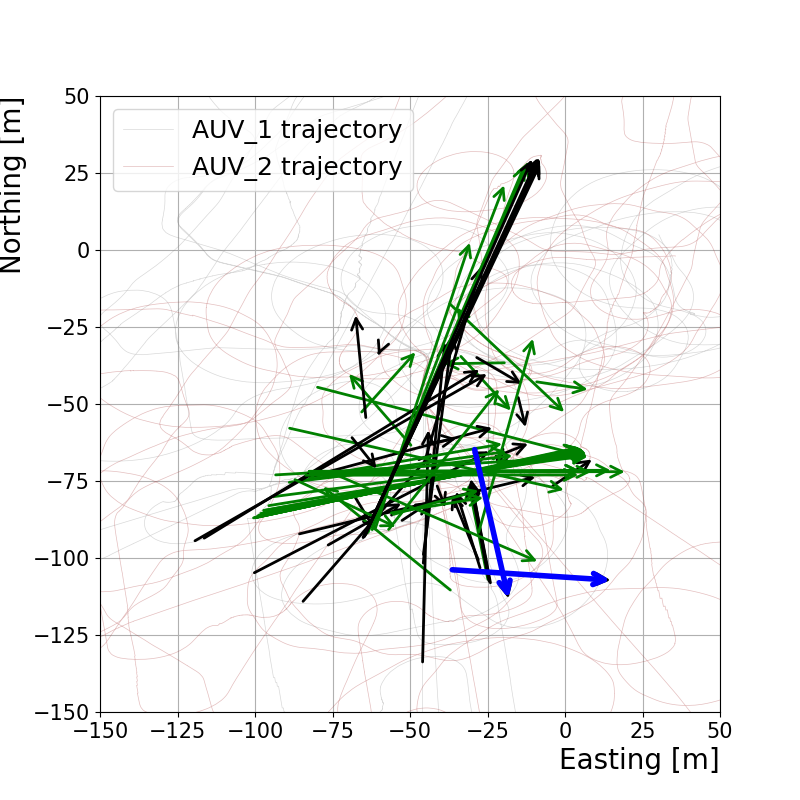}
  \end{subfigure}  

\caption{Two AUVs trajectories (upper figure), Good(red arrows), random (yellow arrows)and bad (blue arrows )inducing points chosen by local agent inside a locality region.}
\label{figure_local_inducing_picked}
\end{figure}


We visualize all training measurements on one locality region (see Figure \ref{figure_local_inducing_picked}) from \textbf{Agent 2 local data set} where AUV 2 acts as the receiving vehicle. Note the locality region $R(r,\textbf{c})$ is not shown in the Figure 2 since it is a ball in $\mathbb{R}^{4}$. The arrows in the figure represent communication events where AUV 1 is broadcasting to AUV 2 so all arrows are pointing from a point along AUV 1 trajectory to a point along the trajectory of AUV 2. Since AUV 2 is the receiving AUV, all of these communication events are only stored onboard AUV 2. Green arrows indicate successful communication events perceived by the agent, while black arrows represent unsuccessful communication events. The agent selects two good/random/bad inducing locations based on the data inside this locality region.
We observe that the selection of two good inducing points that minimizes $tr(\tilde{\mathbf{K}})$ tends to identify locations near data cluster areas, enabling better representation of all data within this region. In contrast, bad inducing locations typically fall in areas that poorly represent the local data set within this region.

Having shown that selected \emph{good} inducing points improve local prediction performance, we extend our analysis to multi-agent contexts. Results demonstrate that sharing \emph{good} inducing points among decentralized agents consistently outperforms sharing \emph{random} or \emph{bad} inducing points. We evaluate performance when agents concatenate received inducing locations and variational distributions to form the decentralized posterior $q(\textbf{f}_{M} | w)$ Section \eqref{inducing_posterior}, then predict on test points across all locality regions.

\begin{figure}[t!]
    \centering
  \begin{subfigure}[b]{0.4\textwidth}
  \includegraphics[width=\textwidth]{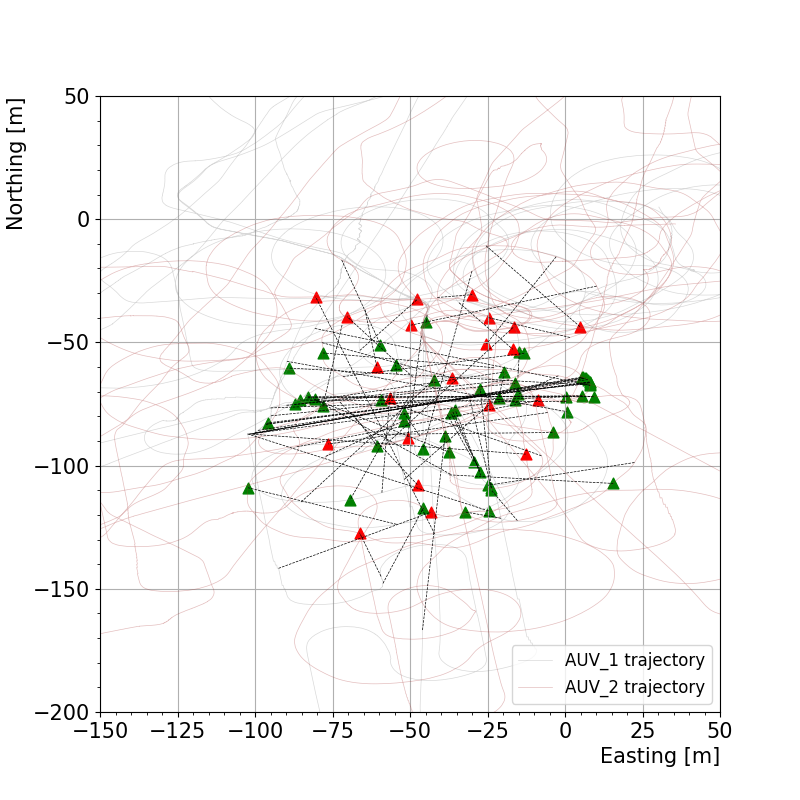}
  \end{subfigure}

  \centering
    \begin{subfigure}[b]{0.4\textwidth}
  \includegraphics[width=\textwidth]{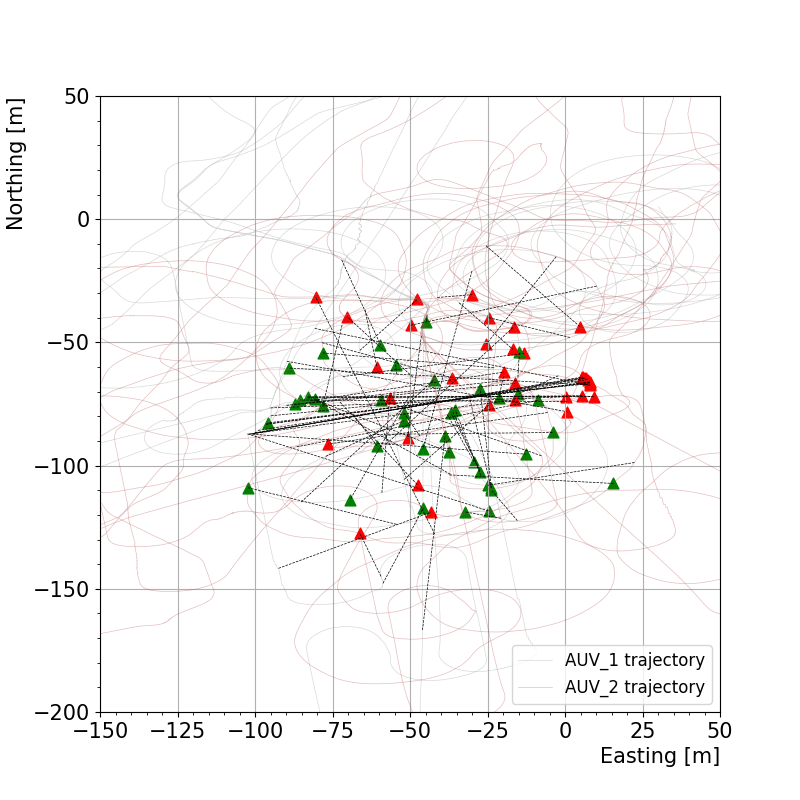}
  \end{subfigure}  

  \centering
    \begin{subfigure}[b]{0.4\textwidth}
  \includegraphics[width=\textwidth]{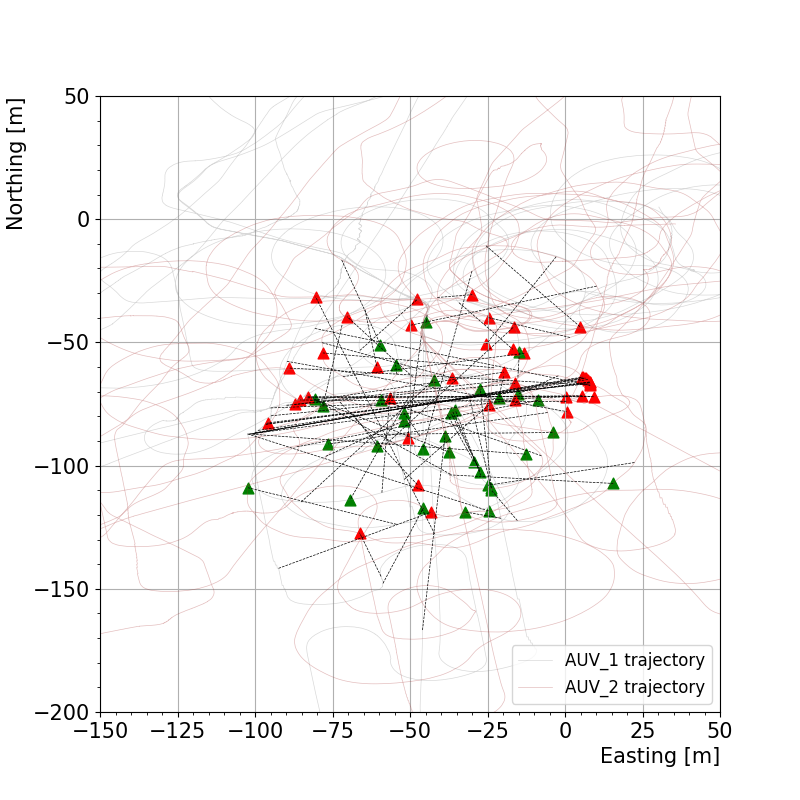}
  \end{subfigure}  
\caption{Two agents decentralized prediction on test points from two different locality regions using good (upper figure)/random (middle figure)/bad (bottom figure)inducing points shared among each agent}
\label{figure_decentralized_inducing_predict}
\end{figure}

Our results clearly demonstrate that in the decentralized scenario, agents benefit from sharing good inducing points with each other under limited communication bandwidth, achieving increased accuracy and decreased negative test likelihood (NLL),  see tables \ref{tab.2agents_1} \ref{tab.2agents_2} and \ref{tab.3agents_1}. We visualize test communication event predictions covered by two locality regions visited by both agents in the \textbf{Dataset 1} (see Figure \ref{figure_decentralized_inducing_predict}). In these figures, green solid triangles indicate correct predictions (predictive probability $p > 0.5$ if $y = 1$ and $p \leq 0.5$ if $y = 0$, while red solid triangles denote incorrect predictions. Although the two locality regions in Figure \ref{figure_decentralized_inducing_predict} are actually partially overlap, our prediction results still demonstrate that sharing good inducing points provides significant advantages over alternative approaches.

\begin{table}[h!] 
\caption{(\textbf{Dataset 1})}
\begin{center} \label{tab.2agents_1}
\begin{tabular}{|c|c|c|c|c|c|} 
\hline
\textbf{SGPC-PG} & \textbf{ACC} & \textbf{NLL}\\
\hline

\hline\hline
\textbf{Good: share one / two} & \textbf{0.6844} / \textbf{0.6941} & \textbf{0.6315} / \textbf{0.5815}\\
\hline
\textbf{Random: share one / two} & 0.5972 / 0.6504 & 0.6674 / 0.6296\\
\hline
\textbf{Bad: share one / two} & 0.5606 / 0.5620 & 0.6765 / 0.6765\\
\hline
\end{tabular}
\end{center}
\end{table}

\begin{table}[h!] 
\caption{(\textbf{Dataset 2})}
\begin{center} \label{tab.2agents_2}
\begin{tabular}{|c|c|c|c|c|c|} 
\hline
\textbf{SGPC-PG} & \textbf{ACC} & \textbf{NLL}\\
\hline

\hline\hline
\textbf{Good: share one / two} & 0.7442 / 0.7442 & \textbf{0.5020} / \textbf{0.4918}\\
\hline
\textbf{Random: share one / two} & 0.7485 / 0.7493 & 0.5756 / 0.5327\\
\hline
\textbf{Bad: share one / two} & 0.7433 / 0.7437 & 0.6320 / 0.6111\\
\hline
\end{tabular}
\end{center}
\end{table}

\begin{table}[h!] 
\caption{(\textbf{Dataset 3})}
\begin{center} \label{tab.3agents_1}
\begin{tabular}{|c|c|c|c|c|c|} 
\hline
\textbf{SGPC-PG} & \textbf{ACC} & \textbf{NLL}\\
\hline

\hline\hline
\textbf{Good: share one / two} & \textbf{0.7101} / \textbf{0.7270} & \textbf{0.6267} / \textbf{0.5976}\\
\hline
\textbf{Random: share one / two} & 0.6979 / 0.6967 & 0.6433 / 0.6273\\
\hline
\textbf{Bad: share one / two} & 0.6947 / 0.6981 & 0.6550 / 0.6532\\
\hline
\end{tabular}
\end{center}
\end{table}

\section{Concluding Remarks}
We present a decentralized data sharing policy that leverages the theoretical advantages of sparse Gaussian process classification with Pólya-Gamma random variables. Our results demonstrate that inducing points selected through this policy offer significant advantages over random and bad inducing points when constructing communication maps under bandwidth limitations. Future research directions include developing more sophisticated sharing policies for scenarios with overlapping locality regions and optimizing the utilization of previously received inducing points to enhance prediction accuracy. These advancements will further improve coordination capabilities among autonomous underwater vehicles operating in challenging communication environments.

\IEEEtriggeratref{10}
\bibliography{sources}
\bibliographystyle{ieeetr}
\end{document}